\documentclass{article}
\usepackage{spconf,amsmath,graphicx}
\usepackage{amssymb}
\usepackage{multirow}
\usepackage{makecell}
\usepackage{lineno}

\usepackage{xcolor}

\title{A NON-HIERARCHICAL ATTENTION NETWORK WITH MODALITY DROPOUT FOR TEXTUAL RESPONSE GENERATION IN MULTIMODAL DIALOGUE SYSTEMS}

\name{Rongyi Sun$^{1}$ \quad Borun Chen$^{1}$ \quad Qingyu Zhou$^{2}$ \quad Yinghui Li$^{1}$\quad  Yunbo Cao$^{2}$ \quad Hai-Tao Zheng$^{1*}$
\thanks{* Corresponding author. (E-mail: zheng.haitao@sz.tsinghua.edu.cn)
}}
\address{$^{1}$Shenzhen International Graduate School, Tsinghua University \\
      $^{2}$Tencent Cloud Xiaowei}

\begin{document}
%
\maketitle
\begin{abstract}
Existing text- and image-based multimodal dialogue systems use the traditional Hierarchical Recurrent Encoder-Decoder (HRED) framework, which has an utterance-level encoder to model utterance representation and a context-level encoder to model context representation. Although pioneer efforts have shown promising performances, they still suffer from the following challenges: (1) the interaction between textual features and visual features is not fine-grained enough. (2) the context representation can not provide a complete representation for the context. To address the issues mentioned above, we propose a non-hierarchical attention network with modality dropout, which abandons the HRED framework and utilizes attention modules to encode each utterance and model the context representation. 
To evaluate our proposed model, we conduct comprehensive experiments on a public multimodal dialogue dataset.
Automatic and human evaluation demonstrate that our proposed model outperforms the existing methods and achieves state-of-the-art performance.

\end{abstract}

\section{Introduction}
\label{sec:intro}
\begin{figure}[t]
 \centering
 \centerline{\includegraphics[width=8.5cm]{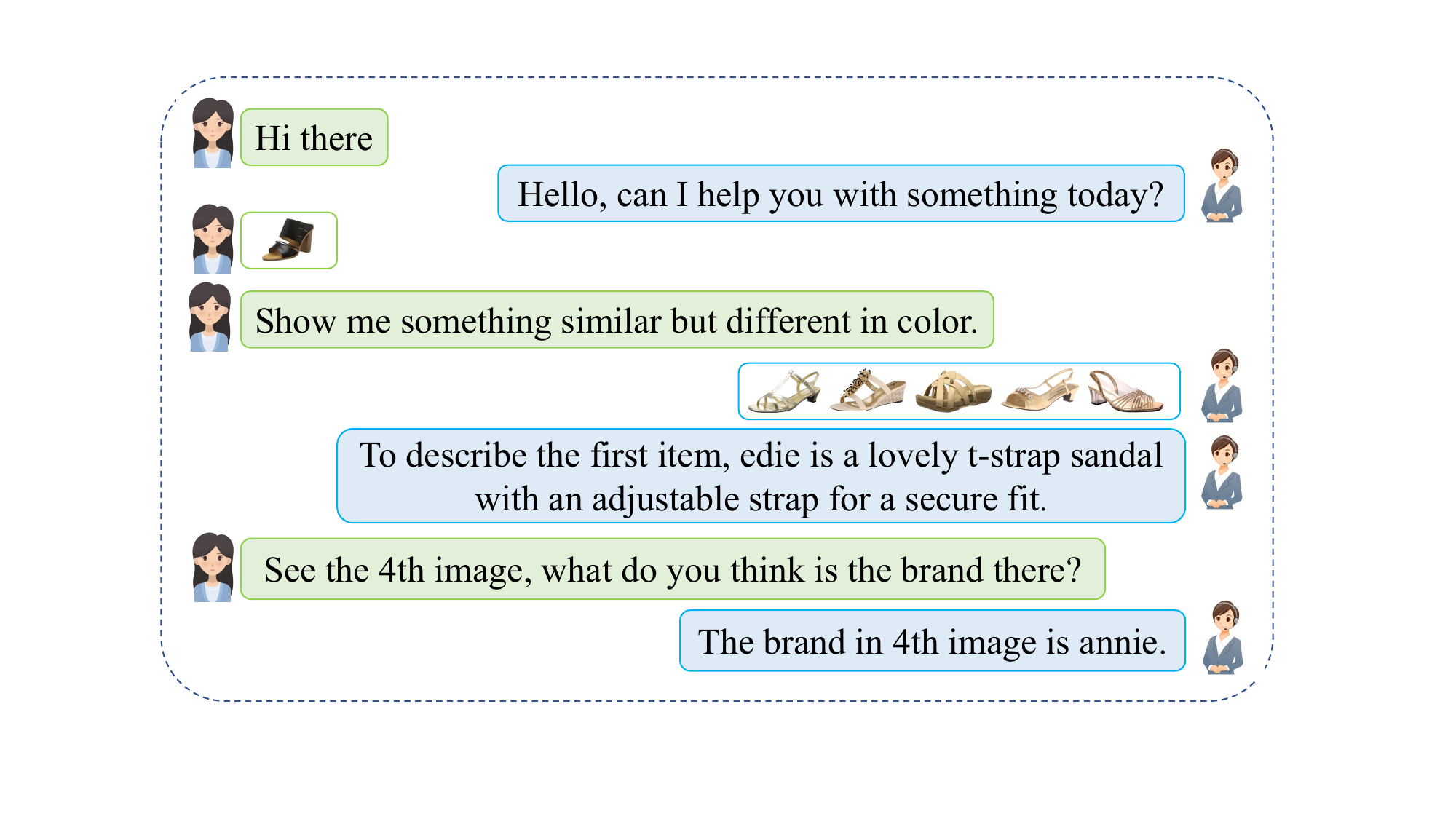}}

\caption{An example of multimodal dialogues between a user and a chatbot.}
\label{fig:example}
\end{figure}

With the development of social media, dialogue systems are no longer confined to the textual form, and pictures are gradually playing an essential role in online conversations \cite{MMD}. As shown in Figure \ref{fig:example}, the customer expresses requirements and questions about products, and the chatbot generates the multimodal responses according to the context, which remarkably demonstrates that ``one picture is worth a thousand words". 
Inspired by this, incorporating images into the traditional textual dialogue systems, i.e., multimodal dialogue systems, has been attracting more and more research interest \cite{MMD,UMD,MAGIC,OAM,sota}.

Most existing methods build multimodal dialogue systems by adding a multimodal feature fusion module to the traditional textual dialogue systems.
Along this line, \cite{MMD,UMD,MAGIC,OAM} take the traditional Hierarchical Recurrent Encoder-Decoder (HRED) \cite{HRED} as the basic framework and propose different fusion modules. 
Although these models have shown promising performances, they still suffer from two problems.
Firstly, the interaction between textual features and visual features is not fine-grained enough. For each utterance in the context, they first compute a sentence embedding vector to represent texts and then get the utterance representation by interacting this sentence embedding vector with image feature vectors. This may lead to some word-related attribute features in the images being ignored because the sentence embedding vector tends to focus more on the global information of texts. 
Secondly, the output of the context-level encoder, i.e., the context representation, may not provide a complete representation for the context. For each utterance, the context-level encoder updates the context representation by encoding the output of the utterance-level encoder. 
This hierarchical architecture makes the context-level encoder not see the complete utterance because the utterance-level encoder encodes the utterance into a feature vector, which leads to a loss of word-level information.
However, we argue that which information is lost should be determined by the context-level encoder rather than the utterance-level encoder.
\begin{figure*}[t]
 \centering
 \centerline{\includegraphics[width=17.5cm,height=7.15cm]{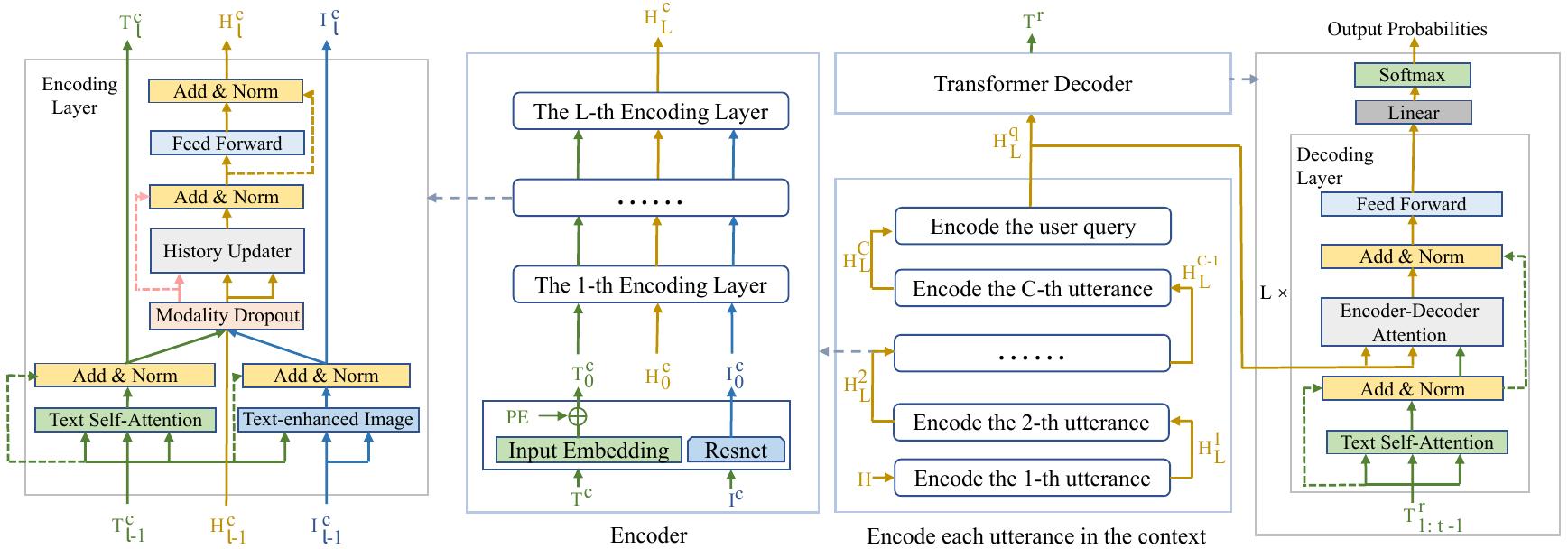}}
\caption{The overall framework of our proposed model, from left to right they are: the inner architecture of a encoding layer, the architecture of the encoder, the entire model, and the architecture of the decoder. The encoder has L sub-layers and PE denotes positional embedding. The decoder generates textual responses according to the output of the encoder.}
\label{fig:model}
\end{figure*}

To address the issues mentioned above, we propose a non-hierarchical attention network with modality dropout.
For the first problem, we use a self-attention module to extract textual features and a text-enhanced image module to extract textual-visual features in each encoding layer.
Both of the modules are based on word level. Thus texts and images are capable of interacting with each other at a finer granularity.
Besides, to fully utilize features and prevent the model from overfitting to one specific modality, we propose a modality dropout strategy to fuse textual and textual-visual features. 
For the second problem, we abandon the hierarchical architecture and design a history updater, which updates context history information after fusing features in each encoding layer instead of the last encoding layer.

We conduct comprehensive experiments on a public multimodal dialogue dataset. 

Experimental results show that our model outperforms state-of-the-art methods on automatic evaluation metrics. 
We also perform human evaluation to verify the quality of generated textual responses. The results indicate that the generated textual responses of our model are fluent and relevant to the context.

\section{Methodology}
\label{sec:method}
\subsection{Problem Formulation}
Given a set of multimodal dialogue utterances between a user and a chatbot $U_C=\{(T^c,I^c)\}_{c=1}^C$ and a user query $(T^q,I^q)$, the task is to generate the system response $(T^r,I^r)$. Here, $T^c=\{w_i^c\}_{i=1}^{m_c}$ denotes the $c$-th text utterance containing $m_c$ words, and $I^c=\{i_j^c\}_{j=1}^{n_c}$ denotes the $c$-th image utterance containing $n_c$ images. Similarly, $T^q=\{w_i^q\}_{i=1}^{m_q}$ contains $m_q$ words and $I^q=\{i_j^q\}_{j=1}^{n_q}$ contains $n_q$ images. 
Mainstream methods  treat textual responses generation and image responses as two separate tasks, and generate the responses by selectively assembling the texts and images manually \cite{MMD,UMD,MAGIC,OAM}. In this paper, we only focus on generating more appropriate textual responses. Therefore, our goal is to maximize the probability of generating the textual response $T^r={\{w_i^r\}_{i=1}^{m_r}}$, which is computed by
\begin{equation}
    p(T^r|U_C;T^q;I^q;\theta) = \prod_{i=1}^{m_r}p(w_i^r|U_C;T^q;I^q;w_{<i}^r;\theta),
    \label{eq:goal}
\end{equation}
where $w_{<i}^r$ denotes generated words and $\theta$ denotes model parameters.

\subsection{Model Architecture}
As shown in Figure \ref{fig:model}, our model has an encoder-decoder architecture like Transformer \cite{transformer:17}. 
In this section, we mainly introduce the encoder, which processes each utterance in the context sequentially and utilizes a history updater to maintain the context history information. 

\subsubsection{Processing Utterance}
For the c-th utterance in $U_C$, we first convert texts $T^c$ into text embeddings $T_0^c \in \mathbb{R}^{m_c \times d_{model}}$ and extract visual features $I_0^c \in \mathbb{R}^{n_c \times d_{model}}$ from images $I^c$ by Resnet \cite{resnet}, where $d_{model}$ is the dimension of word embeddings and visual features. 
The encoder, having $L$ sub-layers, computes the textual features $T_l^c$ by text self-attention module and computes textual-visual features $I_l^c$ by text-enhanced image module. Both of the modules are implemented by multi-head attention mechanism \cite{transformer:17}.
More specifically, at the $l$-th layer, textual features $T_l^c$ are computed by:

\begin{equation}
    \widetilde{T}_l^c = MultiHeadAttention(T_{l-1}^c,T_{l-1}^c,T_{l-1}^c),
\end{equation}
\begin{equation}
    {T}_l^c = LayerNormalization(\widetilde{T}_l^c) + T_{l-1}^c,
\end{equation}    
and textual-visual  $I_l^c$ features are computed by:
\begin{equation}
    \widetilde{I}_l^c = MultiHeadAttention(T_{l-1}^c,I_{l-1}^c,I_{l-1}^c),
\end{equation}
\begin{equation}
    {I}_l^c = LayerNormalization(\widetilde{I}_l^c) + T_{l-1}^c.
\end{equation}
Both of $T_l^c$ and $I_l^c$ are based on word level and thus have more fine-grained information.
To fully utilize these features and prevent the model from overfitting to one specific modality, we propose a modality dropout strategy here to compute the fusion features $M_l^c$. 
We denote the dropout rate in network as $p_{net} \in [0,1]$. Then a random variable $U^l \in [0,1]$ is sampled uniformly for each layer $l$ to influence the calculation process of $M_l^c$:
\begin{equation}
\begin{aligned}
    M_l^c &= \mathbb{I}(U^l<\frac{p_{net}}{2})\cdot{T}_l^c + \mathbb{I}( U^l>1-\frac{p_{net}}{2})\cdot{I}_l^c \\&+ 
    \frac{1}{2} \times \mathbb{I}(\frac{p_{net}}{2} \leq U^l\leq 1-\frac{p_{net}}{2})\cdot  ({T}_l^c + {I}_l^c),
\end{aligned}
\label{eq:drop}
\end{equation} 
where $\mathbb{I}(\cdot)$ is the indicator function. In this way, for each layer, there is a probability $\frac{p_{net}}{2}$ that only textual features $T_l^c$ or textual-visual features $I_l^c$ are used; w.p. $1-p_{net}$, both features are used. 
Note that $p_{net}$ is set to 0 in the inference phase regardless of what $p_{net}$ is during training process.

\subsubsection{History Updater}
Instead of computing context representation after the last encoding layer, we propose a history updater to maintain the context history information $H_l^c$ after fusing features in each encoding layer:
\begin{equation}
    \widetilde{H}_l^c = MultiHeadAttention(M_l^c,H_{l-1}^c,H_{l-1}^c),
\end{equation}
\begin{equation}
    \hat{H}_l^c = LayerNormalization(\widetilde{H}_l^c) + M_l^c,
\end{equation}
\begin{equation}
    H_l^c =  LayerNormalization(FFN(\hat{H}_l^c)) + \hat{H}_l^c,
\end{equation}
where FFN is a fully connected feed-forward network consisting of two linear transformations with a ReLU activation in between \cite{transformer:17}.
As shown in Figure \ref{fig:model}, the context history information $H_l^c$ is integrated throughout the entire encoding process and we finally get $H_L^q$, which will be fed into decoder as the context representation.

In this process, when $c \neq 1 $ and $l = 0 $, the context history information $H_0^c$ comes from the output of the previous time step $H_L^{c-1}$:
\begin{equation}
    H_0^c = H_L^{c-1}.
    \label{eq:history}
\end{equation}
When $c=1$, there is no $H_L^{0}$ to initialize $H_0^1$.
Instead of random initialization, we add a history parameter $H$ and use it to initialize $H_0^{1}$. During the training process, we also optimize the history parameter $H$ to maximize the probability of generating textual responses. Then the Eqn.(\ref{eq:goal}) can be reformulated as:
\begin{equation}
    p(T^r|U_C;T^q;I^q;\theta) = \prod_{i=1}^{m_r}p(w_i^r|U_C;T^q;I^q;w_{<i}^r;\theta;H).
\end{equation}

\section{Experiments}
\subsection{Dataset}
We conduct experiments on a large-scale Multimodal Dialogue (MMD) dataset \cite{MMD}. The MMD dataset includes 
105,439/225,95/225,95 (train/valid/test) conversations between users and chatbots, and each conversation has an average of 40 utterances. Each utterance can be divided into three categories: text-only utterance, image-only utterance, and multimodal utterance. 

\subsection{Evaluation metrics}
Following previous work \cite{UMD,MAGIC,sota}, we use both automatic and human evaluation methods to measure the performance of our proposed model. For automatic metrics, 
we utilize BLEU \cite{BLEU} and NIST \cite{nist} to measure the similarity between the generated and reference responses.
For human evaluation, we randomly select 300 generated responses of our model and corresponding responses of MAGIC to get 300 response pairs. Then we ask three experts to compare these pairs on fluency, relevance, and logical consistency. 
If two responses in a pair are both meaningful or inappropriate, the comparison of this pair is labeled as “tie”, otherwise each response is tagged with a “win” or “lose”. 
Ultimately, we compute the average result of three experts and calculate their kappa scores \cite{kappa} which indicate a moderate agreement among the annotators. 
\begin{table*}[t] 
\centering
\begin{tabular}{m{1.8cm}<{\centering}|m{5cm}<{\centering}|m{1.5cm}<{\centering}|m{1.5cm}<{\centering}|m{1.5cm}<{\centering}|m{1.5cm}<{\centering}|m{1.4cm}<{\centering}}
\hline
\multicolumn{2}{c|}{Methods} & BLEU1 & BLEU2  & BLEU3 &BLEU4  & NIST \\
\hline
\multirow{2}{*}{Text-only} & Seq2Seq$^{\dag}$ \cite{seq2seq} & 35.39 &28.15 & 23.81 & 20.65 & 3.3261 \\
\cline{2-7}
&HRED$^{\dag}$  \cite{HRED} & 35.44 & 26.15 & 20.81 & 17.27 & 3.1007 \\
\hline
\multirow{6}{*}[-0.9ex]{Multimodal} & MHRED$^\dag$  \cite{MMD} & 32.60 & 25.14 & 23.21 & 20.52 & 3.0901 \\
\cline{2-7}
&UMD$^\dag$ \cite{UMD} &44.97 &35.06& 29.22& 25.03& 3.9831 \\
\cline{2-7}
&OAM$^\dag$ \cite{OAM} & 48.30 & 38.24 & 32.03 & 27.42 & 4.3236 \\
\cline{2-7}
&MAGIC \cite{MAGIC} & 51.05 & 40.31 & 33.87 & 28.91 & 4.3162\\
\cline{2-7}
&MATE$^\dag$  \cite{sota} & 56.55 & 47.89 & 42.48 & 38.06 & \textbf{6.0604} \\
\cline{2-7}
& Our & \textbf{62.80} & \textbf{54.88} & \textbf{48.77} & \textbf{43.03} & 5.6495\\
\hline

\end{tabular}
\caption{Objective performance comparison between our proposed model and baselines. $\dag$ denotes the results reported by \cite{sota}. }
\label{tab:objective_result}
\end{table*}

\begin{table*}[t]
\centering

\begin{tabular}{p{2.85cm}|p{0.7cm}p{0.7cm}p{0.7cm}p{0.8cm}|p{0.7cm}p{0.7cm}p{0.7cm}p{0.8cm}|p{0.7cm}p{0.7cm}p{0.7cm}p{0.8cm}}

\hline
& \multicolumn{4}{c|}{Fluency}& \multicolumn{4}{c|}{Relevance} & \multicolumn{4}{c}{Logical Consistency}\\
\hline
Opponent& Win & Loss & Tie & Kappa & Win & Loss & Tie & Kappa & Win & Loss & Tie & Kappa  \\
\hline
Our vs. MAGIC  & 35.3\%& 5.9\% & 58.8\% & ~~0.51 & 51.6\%& 3.8\%& 44.6\%& ~~0.74 & 52.5\% & 5.3\%& 42.2\%& ~~0.77 \\
\hline
MATE vs. MAGIC$^{\dag}$ & 25.8\% & 15.2\% & 59.0\% & ~~0.65 & 48.8\% & 21.3\% & 29.9\% & ~~0.53 & 49.6\%& 15.6\%& 34.8\%& ~~0.51\\
\hline
\end{tabular}
\caption{Human evaluation results. $\dag$ denotes the evaluation results reported by \cite{sota}.}
\label{tab:subjective_result}
\end{table*}
 
\subsection{Baselines and Hyper-parameters}
To evaluate our proposed model, we compare it with the following baselines: Seq2Seq \cite{seq2seq}, HRED \cite{HRED}, MHRED \cite{MMD}, UMD \cite{UMD}, MAGIC \cite{MAGIC}, OAM \cite{OAM} and MATE \cite{sota}.
Following previous work, we take Resnet18 \cite{resnet} as a visual feature extractor and resize input images into $(3,224,224)$ before feeding them into Resnet. The context size is 2. For each utterance in context, the max number of images is 4 and the max length of sentence is 32.
For our model, it has 2 sub layers in encoder and decoder. The dimension of word embedding is 512. The number of heads for multi-head attention is 8. The word embedding parameters are shared between encoder and decoder. We optimize the model with Adam \cite{kingma2014adam} method that the initial learning rate is 0.0004. We use 4 Telsa-V100 GPUs to train 10 epochs with 150 batch size.

\subsection{Results}
\subsubsection{Automatic Evaluation and Human Evaluation}
As shown in Table \ref{tab:objective_result}, our proposed model outperforms all baselines on BLEU. For example, it improves the BLEU-4 score by 14.12 and 4.97 compared with the state-of-the-art models MAGIC and MATE.
Although it does not surpass MATE in terms of NIST (5.6495 vs. 6.0604), it is still far ahead of other baselines.
This superior performance demonstrates the effectiveness of the novel architecture.
As for why our model does not perform as well as MATE in terms of NIST, we believe this is because NIST focuses more on some keywords that occur less frequently, while MATE uses external knowledge to help the generation of keywords.

Table \ref{tab:subjective_result} presents the human evaluation results. 
Since the source code of MATE is not released, we can not compare MATE with our model directly. Therefore, we do this indirectly by comparing our model with MAGIC.
As shown in Table \ref{tab:subjective_result}, 
our model shows a greater advantage than MATE when taking MAGIC as a reference. This demonstrates that our model is capable of generating more natural responses. 
\begin{table}[t]
\centering
\begin{tabular}{p{1cm}|m{1cm}<{\centering}|m{1cm}<{\centering}|m{1cm}<{\centering}|m{1cm}<{\centering}|m{0.9cm}<{\centering}}
\hline
Setting & BLEU1 & BLEU2 & BLEU3 & BLEU4 & NIST \\
\hline
p = 0.0 & 61.19 & 53.22 & 47.03 & 41.30 & 5.4516\\
\hline
p = 0.2 & 61.63 & 53.64 & 47.45 & 41.67 & 5.4517\\
\hline
p = 0.4 & 62.80 & 54.88 & 48.77 & 43.03 & 5.6495\\
\hline
p = 0.6 & 61.41 & 53.42 & 47.26 & 41.50 & 5.4268\\
\hline
p = 0.8 & 60.73 & 52.72 & 46.49 & 40.73 & 5.4168\\
\hline
p = 1.0 & 60.22 & 52.28 & 46.13 & 40.44 & 5.3761\\
\hline
\end{tabular}
\caption{Different $p_{net}$ for modality dropout.}
\label{tab:abalation_study}
\end{table}

\subsubsection{Ablation Study}
To further verify the effectiveness of the modality dropout and the history updater, we design the corresponding ablation studies. \textbf{Firstly}, we investigate the effect of different $p_{net}$ on performance.
As shown in Table \ref{tab:abalation_study}, we set $p_{net} \in \{0.0,0.2,0.4,0.6,0.8,1.0\}$ and we can see that the performance of the model shows a trend of rising at first and then falling as $p_{net}$ increases.
Especially when $p_{net} = 0.4$, the model has the best performance. This result reflects the improvement brought by the modality dropout.
\textbf{Secondly}, we prove the history updater is effective. When $p_{net}$ equals to 0, the modality dropout strategy is not working and only the history updater is working.
Compared with MATE, which uses Transformer and concatenates the features of each utterance as a context representation, we can see our model performs better from Table \ref{tab:history}.
\textbf{Thirdly}, we set $p_{net} = 0.4$ and do not optimize the history parameter $H$ in the training process. As we can see, the performance dramatically drops when we choose to fix the history parameter $H$ after random initialization, which demonstrates that the history parameter is capable of helping the model start a new conversation.

\begin{table}[t]
\centering
\begin{tabular}{p{1cm}|m{1cm}<{\centering}|m{1cm}<{\centering}|m{1cm}<{\centering}|m{1cm}<{\centering}|m{0.9cm}<{\centering}}
\hline
Setting & BLEU1 & BLEU2 & BLEU3 & BLEU4 & NIST  \\
\hline
MATE & 56.55 & 47.89 & 42.48 & 38.06 & 6.0604\\
\hline
p = 0.0 & 61.19 & 53.22 & 47.03 & 41.30 & 5.4516\\
\hline
\end{tabular}
\caption{The effect of history updater. }
\label{tab:history}
\end{table}

\begin{table}[t]
\centering
\begin{tabular}{p{0.9cm}|m{1cm}<{\centering}|m{1cm}<{\centering}|m{1cm}<{\centering}|m{1cm}<{\centering}|m{0.9cm}<{\centering}}
\hline
Setting & BLEU1 & BLEU2 & BLEU3 & BLEU4 & NIST  \\
\hline
fixed & 59.95 & 52.53 & 46.78 & 41.37 & 5.3599\\
\hline
training & 62.80 & 54.88 & 48.77 & 43.03 & 5.6495 \\
\hline
\end{tabular}
\caption{Different strategies are applied to $H$. }
\label{tab:wo_history}
\end{table}

\section{Conclusion}

In this paper, we propose a non-hierarchical attention network with modality dropout strategy for textual responses generation in multimodal dialogue systems, which allows for finer-grained interaction between multimodal features, and encodes contextual representations without information loss.
Experimental results demonstrates that our proposed model brings a leap in performance to multimodal dialogue systems.
\vfill\pagebreak

\bibliographystyle{IEEEbib}
\bibliography{main}

\end{document}